# Multi-Class Lane Semantic Segmentation using Efficient Convolutional Networks

Shao-Yuan Lo, Hsueh-Ming Hang
National Chiao Tung University
sylo95.eecs02@g2.nctu.edu.tw, hmhang@nctu.edu.tw

Sheng-Wei Chan, Jing-Jhih Lin
Industrial Technology Research Institute
{ShengWeiChan, jeromelin}@itri.org.tw

*Abstract*—Lane detection plays an important role in a self-driving vehicle. Several studies leverage a semantic segmentation network to extract robust lane features, but few of them can distinguish different types of lanes. In this paper, we focus on the problem of multi-class lane semantic segmentation. Based on the observation that the lane is a small-size and narrow-width object in a road scene image, we propose two techniques, Feature Size Selection (FSS) and Degressive Dilation Block (DD Block). The FSS allows a network to extract thin lane features using appropriate feature sizes. To acquire fine-grained spatial information, the DD Block is made of a series of dilated convolutions with degressive dilation rates. Experimental results show that the proposed techniques provide obvious improvement in accuracy, while they achieve the same or faster inference speed compared to the baseline system, and can run at real-time on high-resolution images.

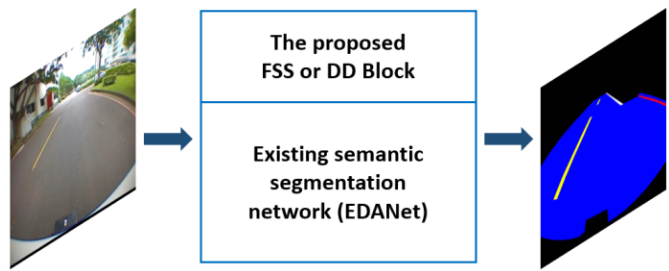

Figure 1. The proposed techniques applied to an existing semantic segmentation network for multi-class lane semantic segmentation.

## I. Introduction

Lane detection is a paramount technology in road scene understanding for autonomous driving. In general, a lane detection algorithm includes three steps: lane feature extraction, feature segment grouping, and lane model fitting [5]. Among these steps, the first step is the most critical and challenging one, so we focus on this step in this work. Traditional lane feature extraction approaches need carefully designed image processing procedures to acquire lane features [1]. For example, gradient-based methods calculate the gradient to capture lane boundaries [13,14,19]. Researchers further combine multiple sources of information, including intensity, color, and edge [17] to enhance detection capability. Lane shape models establish hypothetical criteria to detect lanes, such as hyperbola pair [16] and B-snake [18]. However, these conventional algorithms are not robust enough to resist the impact of environmental variations, such as noises, illumination changes, and weather conditions.

In recent years, deep convolutional neural networks (CNNs) made a breakthrough in the field of computer vision [7,10,12]. Some researchers adopt semantic segmentation networks for lane feature extraction. These CNN-based systems have much higher robustness and are less sensitive to environmental variations. Nevertheless, to the best of our knowledge, few existing studies focus on a challenging task, "multi-class lane semantic segmentation" (see Fig. 1). For instance, LMD [6] designs a network architecture based on SegNet [2] to segment lanes. LMD achieves high performance with real-time inference speed, but it does not distinguish different types of lanes. Zang *et al.* [21] used a CNN to find two-class lane pixels, yellow lines and white lines, yet their method can only process very small size, 32×32, images. Recognizing different types of lanes, such as yellow line, red line, double line, solid line, and dashed line, is critical for a self-driving vehicle since it should understand the meaning of these road lane markings. Nevertheless, the well-known DeepLabv3+ [4], one of the most top-performing segmentation models, and ICNet [22], which aims at self-driving applications have not paid attention to segment multi-class lanes. Based on our experiences, without appropriate modifications, these mainstream architectures, even though powerful, may not be suitable to undertake the task of multi-class lane semantic segmentation.

In this paper, we propose two techniques, Feature Size Selection (FSS) and Degressive Dilation Block (DD Block), to modify an existing semantic segmentation network. The ideas come from our observation that the lane marking is a small-size and narrow-width object in a typical road scene image. EDANet [11] has a good balance between accuracy and inference speed, and it is a proper system for autonomous driving. As a result, we choose it as the baseline architecture and apply our techniques on it (see Fig. 1).

EDANet starts with two Downsampling Blocks to extract features. However, the downsampling process tends to lose detailed spatial information, which is especially detrimental to small objects. Some thin lanes might even be discarded entirely. On the other hand, if we remove the downsampling operations, the receptive field of the network would shrink. Therefore, we extract features at different feature map sizes to investigate the best use of the downsampling operation for lane segmentation. We call this strategy as Feature Size Selection (FSS). Next, similar to many other networks, EDANet employs dilated convolutions with incremental dilation rates to enlarge the receptive field progressively. Still, owing to the sparsity of convolutional kernels, the dilated convolution skips some spatial pixel samples and cannot aggregate complete and detailed information of small objects. Hamaguchi *et al.* [8] developed the LFE module, consisting of several convolutional layers with degressive dilation rates, to extract local features. We adopt the basic concept of the LFE module but use one EDA module as a unit instead of one convolutional layer. We name the stacked EDA modules with degressive dilation rates as Degressive Dilation Block (DD Block).

In summary, this work explores a challenging task, multi-class lane semantic segmentation. We extend the current semantic segmentation system to being able to distinguish various types of lane markings. We evaluate our system on ITRI dataset, which is created by Industrial Technology Research Institute (ITRI). The two proposed techniques make clear improvement on accuracy and achieve the same or higher inference speed compared to the baseline network at the same time. They can run at real-time on high-resolution images and thus applicable to the autonomous driving applications.

## II. METHOD

The lane is a relatively small and thin object in a road scene image, so we need to develop new strategies focusing on small object semantic segmentation. In this section, we introduce the details of the two proposed techniques, FSS and DD Block. These two methods are constructed on the baseline network, EDANet [11], and we name them as EDA-FSS and EDA-DDB, respectively.

### A. Feature Size Selection

The detailed spatial information is important for accurate lane localization. Typical CNNs have several downsampling layers, where EDANet has three downsampling operations. It starts extracting features after two Downsampling Blocks, i.e., on the feature maps whose sizes are 1/4 to the input size. Fig. 2(a) illustrates the architecture of EDANet. The downsampled feature maps tend to lose precise boundary information, and sometimes the thin lanes would vanish. However, extracting features on large feature maps requires more computations. In addition, it is more difficult to obtain wide enough receptive fields. Making a good balance between these two needs is a challenge.

EDA-FSS, the proposed architecture, is shown in Fig. 2(b). It places an additional EDA Block 0, which consists of two EDA modules, between the first and the second Downsampling Block in EDANet. With EDA Block 0, EDA-FSS is enabled to extract features on larger feature maps (1/2 to the input size), and thus small objects or detailed boundaries can be detected. In order to maintain similar computational complexity for high inference speed and fair comparison, we reduce the number of EDA modules in EDA Block 1 and EDA Block 2 from 5 to 4 and 8 to 5, respectively. The growth rate of each convolutional layer is also decreased from 40 to 30. The overall depth and width of EDA-FSS are shallower and narrower than those of EDANet since performing operations at an early stage needs more computations. For this specific task, this trade-off improves segmentation accuracy successfully.

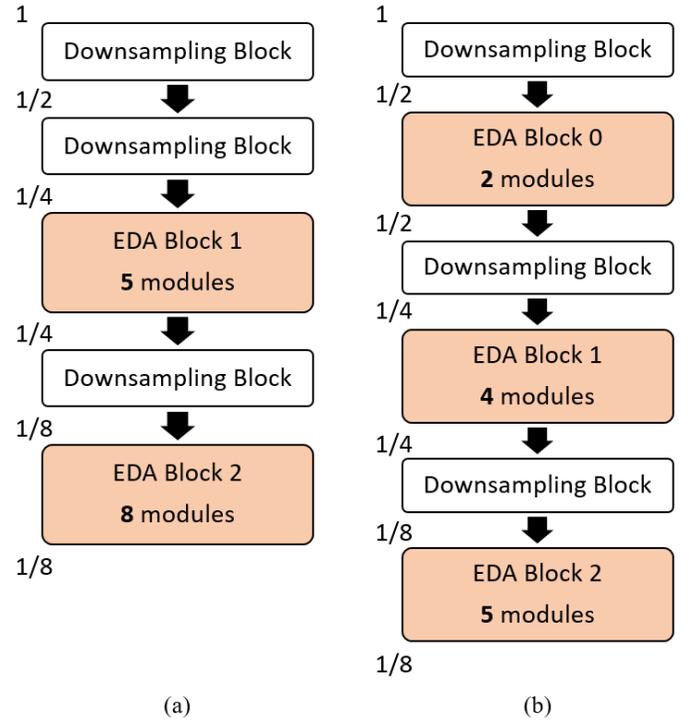

Figure 2. (a) EDANet [11]. (b) The proposed EDA-FSS. The numbers next to each block denote the feature map size ratios to the input image size.

### B. Degressive Dilation Block

The dilated convolution is widely used by a lot of semantic segmentation networks [3,20]. Typically, these architectures leverage the dilated convolution with incremental dilation rates to enlarge the receptive field in a gradual manner. EDANet adopts the same concept, too. However, this design suffers from the problem that the filter supports of adjacent pixels might produce inconsistent spatial information in local features. In other words, the receptive fields of adjacent pixels in deeper layers probably only slightly overlap with each other in shallower layers. In addition, it may fail to gather the essential local features because of the sparsity of convolutional kernels. This problem would cause inaccurate recognition of small and thin objects. Hamaguchi *et al.* [8] proposed an architecture called Local Feature Extraction (LFE) module, which aims at the task of segmenting small instances in remote sensing imagery. The LFE module is composed of several convolutional layers with degressive dilation rates, and it can solve the aforementioned problems to a certain extent.

Inspired by LFE module, we propose a new structure, DD Block. Unlike LFE module using one convolutional layer as a structural unit, its unit is one EDA module. DD Block consists of four EDA modules with the degressive dilation rates: 8, 4, 2, and 1. We insert the proposed DD Block into EDANet to construct EDA-DDB. Table I compares the structure between EDANet and EDA-DDB. EDA-DDB adds a DD Block and reduces the number of EDA modules in EDA Block 2 from eight to four. The dilation rates are 2, 4, 8, and 16, in sequence. EDA-DDB has the same number of parameters and computational complexity as EDANet.

TABLE I. Description of EDANet [11] and the proposed EDA-DDB structures. The numbers denote the dilation rates of each EDA module. Only EDA Block 2 and DD Block are shown since the rest of these two networks are identical.

| Network | EDANet [11] | EDA-DDB |
|---|---|---|
| EDA Block 2 | 2 | 2 |
|  | 2 | 4 |
|  | 4 | 8 |
|  | 4 | 16 |
|  | 8 | - |
|  | 8 | - |
|  | 16 | - |
|  | 16 | - |
| DD Block | - | 8 |
|  | - | 4 |
|  | - | 2 |
|  | - | 1 |

III. EXPERIMENTS

We construct several variants of the proposed FSS and DD Block. Then, we conduct a series of experiments to analyze their performance. In this section, we first introduce the dataset we use. Next, the implementation details are described. Finally, the experimental results are reported.

A. ITRI Dataset

The ITRI dataset was created by the Mechanical and Mechatronics Systems Research Lab, Industrial Technology Research Institute (ITRI), Taiwan. To the best of our knowledge, it is one of the first datasets having pixel-wise annotations of different types of traffic lanes. This dataset is still growing, so we only use its currently available subsets. We combine the given sets C2, C3, C7, C13, C18, and C20 to form a training set with 2,192 images. The test set consists of sets C4 and C14 with 567 images in total. The dataset has totally six classes, including four types of traffic lanes, road, and the undefined class. The four types of the lane are double solid yellow line, single dashed yellow line, single solid red line, and single solid white line. All the other marks on roads are labeled as the road class. The image resolution is 480×720. Fig. 3 shows some samples.

B. Implementation Details

We follow the training setup similar to that in EDANet [11]. Our networks are trained by using Adam optimization [9]. The weight decay is set to $1e^{-4}$, and the batch size is 16. We set the initial learning rate to $5e^{-4}$ and adopt the poly learning rate policy; that is, the learning rate is multiplied by $(1 - iter/\max\_iter)^{power}$ with power 0.9. Random horizontal flip and a translation of 0~2 pixels on both axes are adopted for data augmentation. We adopt the mean of intersection-over-union (mIoU) as a metric for accuracy evaluation. We do not use any testing tricks, e.g., multi-crop and multi-scale testing. Our computing device is a single GTX 1080Ti.

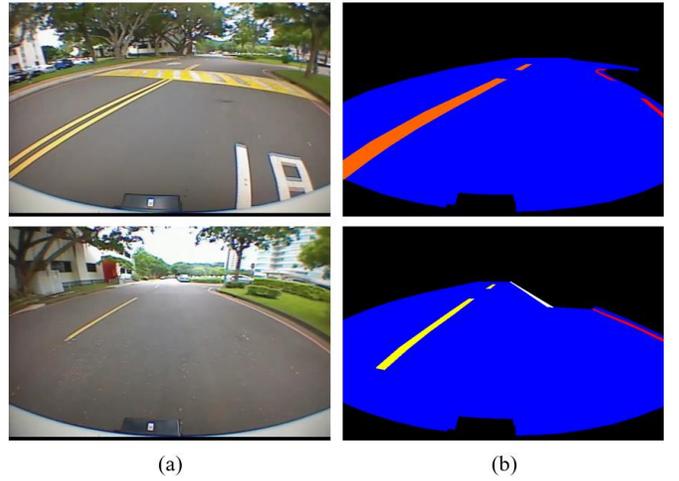

Figure 3. Samples of ITRI dataset. (a) RGB image. (b) Ground truth map.

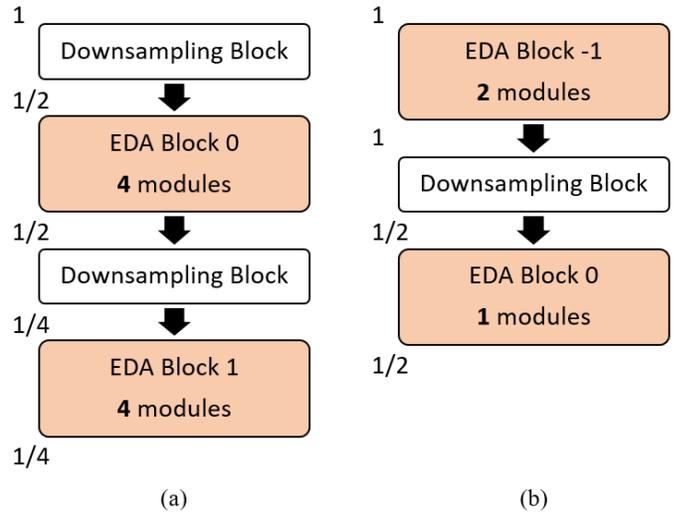

Figure 4. (a) Network A. (b) Network B. The numbers next to each block denote the feature map size ratios to the input image size.

C. Feature Size Selection

We evaluate the performance of the proposed EDA-FSS on the ITRI dataset. We also design and evaluate two network variants of EDA-FSS called Network A and Network B for comparison. Their architectures are shown in Fig. 4. Compared to EDA-FSS, Network A increases the number of EDA modules in EDA Block 0 from 2 to 4, but the entire EDA Block 2 is discarded to maintain similar computational complexity. Network B further adds EDA Block -1 in, which consists of two EDA modules, in front of the first Downsampling Block. This block extract features on the original image size (480×720). Again, because of the consideration for computational cost and fair comparison, the number of EDA modules in EDA Block 0 is set to 1. Furthermore, the entire EDA Block 1 is removed from Network B.

Table II reports the experimental results. From EDANet to Network B, the feature map sizes become larger, but their network depths become shallower in order to keep similar computational complexity. Then, we explore the trade-off between feature size and network depth. First, we can see that our baseline, EDANet, outperforms another famous efficient segmentation network, ERFNet [15], on both mIoU accuracy and run time. Second, the proposed EDA-FSS surpasses EDANet in accuracy and achieves even shorter run time. Therefore, we conclude that placing some convolutional layers at the early stage is able to extract more useful features for small object segmentation since more spatial information is retained. On the other hand, the relatively poor performance of Network A and Network B indicates that very shallow structures are not a good solution though they extract features on larger sizes. They fail to obtain sufficiently wide receptive fields, and their computational cost is even higher. Apparently, EDA-FSS achieves a better balance. Fig. 5 compares their visual results. We can observe that ERFNet fails to recognize the double solid yellow lane as one united lane in the first sample image. EDANet has the same problem in the second sample image. Next, Network B produces holes on the road in the second sample image. In general, EDA-FSS outputs the most precise results in which the patterns are intact and correctly detected.

TABLE II. Evaluation results of the experiment on FSS. The IoU scores of each class except the undefined class are listed. DS-Y: Double solid yellow lane. SD-Y: Single dashed yellow lane. SS-R: Single solid red lane. SS-W: Single solid white lane.

| Network | DS-Y | SD-Y | SS-R | SS-W | Road | mIoU | Run time |
|---|---|---|---|---|---|---|---|
| ERFNet [15] | 78.6 | 82.4 | 39.1 | 46.1 | 96.9 | 73.3 | 24ms |
| EDANet [11] | 85.2 | 69.4 | 38.4 | 61.4 | 97.0 | 74.7 | 9.2ms |
| Network A | 85.1 | 72.1 | 37.7 | 57.1 | 96.9 | 74.2 | 12ms |
| Network B | 81.8 | 65.9 | 41.7 | 57.1 | 96.2 | 73.1 | 18ms |
| EDA-FSS | 84.5 | 74.4 | 43.0 | 54.8 | 96.9 | **75.0** | **8.7ms** |

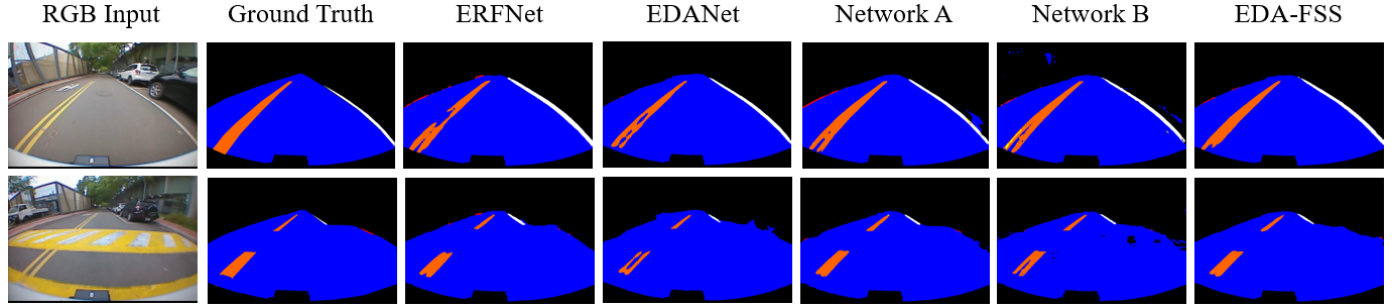

Figure 5. Sample visual results produced by ERFNet [15], EDANet [11], Network A, Network B, and the proposed EDA-FSS.

TABLE III. Description of the structures of each variant. The numbers denote the dilation rates of each EDA module. Only EDA Block 2 and extra block are shown since the rest of these networks are identical.

| Network | EDA-w/o-di | EDA-DDB-L | EDA-Large-1 | EDA-Large-16 |
|---|---|---|---|---|
| EDA Block 2 | 1 | 2 | 2 | 2 |
| | 1 | 2 | 2 | 2 |
| | 1 | 4 | 4 | 4 |
| | 1 | 4 | 4 | 4 |
| | 1 | 8 | 8 | 8 |
| | 1 | 8 | 8 | 8 |
| | 1 | 16 | 16 | 16 |
| | 1 | 16 | 16 | 16 |
| Extra Block | - | 8 | 1 | 16 |
| | - | 4 | 1 | 16 |
| | - | 2 | 1 | 16 |
| | - | 1 | 1 | 16 |

### D. Degressive Dilation Block

We evaluate the performance of the proposed EDA-DDB. Likewise, we design several variants for ablation study. Table III compares the structures of each variant. EDA-w/o-di does not include any dilated convolution. It is used to assess the effectiveness of using dilated convolution. EDA-DDB-L attaches a DD Block to EDANet directly without reducing the number of EDA modules in EDA Block 2. EDA-Large-1 adds four EDA modules with a fixed dilation rate 1. This extra block has the same number of parameter and complexity as DD Block. EDA-Large-16 is nearly identical to EDA-Large-1 except that the fixed dilation rate of its extra block is 16. EDA-Large-1 and EDA-Large-16 are used to check whether the improvement brought from the DD Block in EDA-DDB-L is due to the degressive dilation design or merely the additional parameters.

As shown in Table IV, EDA-w/o-di is less accurate than EDANet, which proves the effectiveness of dilated convolution. Next, EDA-DDB-L performs better than EDANet but worse than EDA-Large-16. As a result, the improvement of EDA-DDB-L may be due to the increased parameters rather than the notion of DD Block. Finally, the proposed EDA-DDB achieves the best performance, which indicates leveraging the decreasing dilation rates in a proper way is still able to lead to improvements. The lane is a thin object, so it would not benefit from a too deep structure. This is one of the reasons that EDA-DDB can surpass EDA-DDB-L though it has fewer parameters. In summary, EDA-DDB successfully makes noticeable better performance than the baseline with the same network complexity. Moreover, it maintains a very short run time even on high-resolution images (480×720). Fig. 6 demonstrates several visual results produced by EDA-DDB. Basically, it is able to detect most of the lanes in the road scene images and can correctly distinguish different types of lanes. Still, it sometimes misses side lanes, especially the red lanes on the right-hand side. This is what we can further investigate.

TABLE IV. Evaluation results of the experiment on DD Block. The IoU scores of each class except the undefined class are listed. DS-Y: Double solid yellow lane. SD-Y: Single dashed yellow lane. SS-R: Single solid red lane. SS-W: Single solid white lane.

| Network | DS-Y | SD-Y | SS-R | SS-W | Road | mIoU | Run time |
|---|---|---|---|---|---|---|---|
| EDANet [11] | 85.2 | 69.4 | 38.4 | 61.4 | 97.0 | 74.7 | **9.2ms** |
| EDA-w/o-di | 78.8 | 70.9 | 40.2 | 60.9 | 96.7 | 74.0 | **9.2ms** |
| EDA-DDB-L | 86.3 | 71.9 | 40.9 | 59.1 | 97.1 | 75.3 | 12ms |
| EDA-Large-1 | 82.7 | 68.7 | 37.4 | 58.7 | 96.6 | 73.4 | 12ms |
| EDA-Large-16 | 81.1 | 71.4 | 44.0 | 62.5 | 96.8 | 75.4 | 12ms |
| EDA-DDB | 86.6 | 76.4 | 43.6 | 57.6 | 97.1 | **75.9** | **9.2ms** |

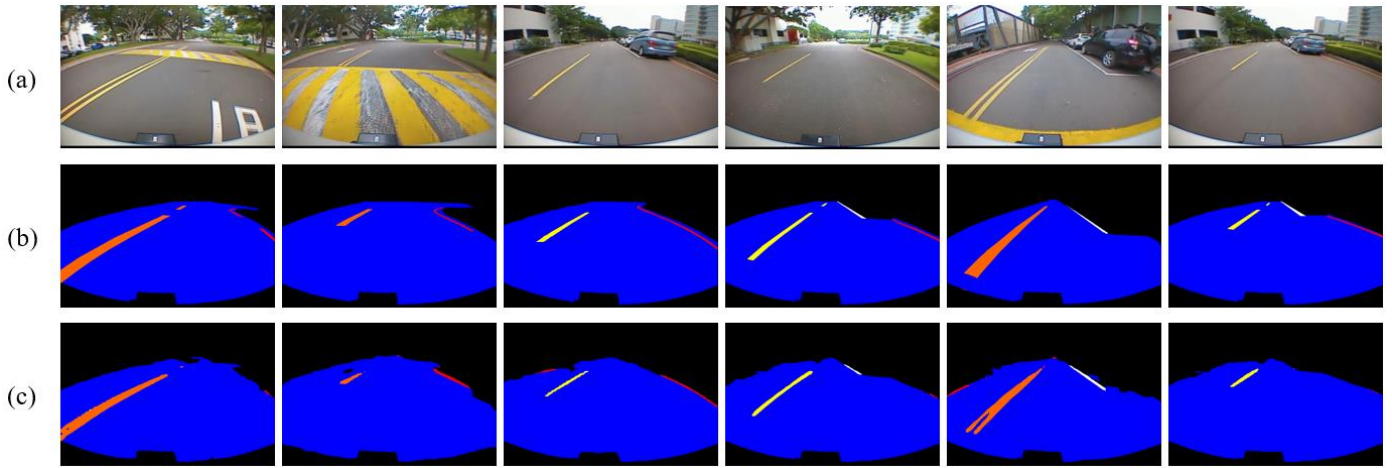

Figure 6. Sample visual results produced by the proposed EDA-DDB. (a) RGB input. (b) Ground truth. (c) EDA-DBB.

## IV. CONCLUSION

In this paper, we propose two techniques, feature size selection and DD Block, for multi-class lane semantic segmentation. We find that using larger feature sizes can acquire more localization information for small object segmentation, but keeping a good balance between the network depth and inference speed is critical. Next, the proposed EDA-DDB, which includes a DD Block in the modified EDANet, achieves obvious improvement in accuracy by more fine-grained spatial information. Our systems generate robust lane features, which can be used easily by lane marking post-processing algorithms.

Moreover, they are able to run at real-time on high-resolution inputs, so they are feasible for real self-driving cars. The techniques proposed in this work are not limited to the use of lane detection, but could also be tried at other small object semantic segmentation tasks.

ACKNOWLEDGMENT

This work was supported in part by the Mechanical and Mechatronics Systems Research Lab., ITRI, under Contract 3000547822.